\documentclass{article}
\usepackage{spconf,amsmath,graphicx}

\usepackage{algorithm}
\usepackage{algpseudocode}

\usepackage{ctable}
\usepackage{mathpazo} 

\title{Fast threshold optimization for multi-label audio tagging using Surrogate gradient learning}
\name{Thomas Pellegrini$^{1}$, Timoth\'ee Masquelier$^2$}
\address{$^{1}$IRIT, Universit\'e Paul Sabatier, CNRS, Toulouse, France\\
	$^2$CERCO UMR 5549, CNRS -- Universit\'e Toulouse 3, Toulouse, France}
\begin{document}
%
\maketitle
\begin{abstract}
Multi-label audio tagging consists of assigning sets of tags to audio recordings. At inference time, thresholds are applied on the confidence scores outputted by a probabilistic classifier, in order to decide which classes are detected active. In this work, we consider having at disposal a trained classifier and we seek to automatically optimize the decision thresholds according to a performance metric of interest, in our case F-measure (micro-F1). We propose a new method, called SGL\-Thresh for Surrogate Gradient Learning of Thresholds, that makes use of gradient descent. Since F1 is not differentiable, we propose to approximate the thresholding operation gradients with the gradients of a sigmoid function. We report experiments on three datasets, using state-of-the-art pre-trained deep neural networks. In all cases, SGL\-Thresh outperformed three other approaches: a default threshold value (defThresh), an heuristic search algorithm and a method estimating F1 gradients numerically. It reached 54.9\% F1 on AudioSet eval, compared to 50.7\% with defThresh. SGL\-Thresh is very fast and scalable to a large number of tags\footnote{To facilitate reproducibility, data and source code in Pytorch are available online: https://github.com/topel/SGL\-Thresh}.
\end{abstract}
\begin{keywords}
Audio tagging, Surrogate Gradient Learning, Automatic threshold optimization.
\end{keywords}
\section{Introduction}
\label{sec:intro}

In this work, we are interested in sound event detection (SED), more precisely in audio tagging (AT). AT consists in automatically assigning sets of binary labels, called tags, to real-life recordings. Audio events are often polyphonic, which means that several types of events may occur simultaneously, \textit{e.g.}, cat meowing and speech. Thus, AT is a multi-label classification (MLC) task. AT and SED in general --- SED consists in localizing the onset and offset of events --- have gained a lot of interest within the audio content analysis community this last decade, because of deep neural networks (DNNs) and the availability of large-scale datasets such as Audioset~\cite{audioset}. Stimulating challenges have also been organized, \textit{e.g.}, the Detection and Classification of Acoustic Scenes and Events (DCASE) challenges\footnote{http://dcase.community/}, in which AT is key in several tasks.

State-of-the-art AT systems rely on DNNs, typically convolutional neural networks (CNN) and  convolutional recurrent neural networks~\cite{virtanen2018computational,kong2019panns}. Some of these models are open-source and available to the research community, both their architectures and their trained weights~\cite{kong2019panns}. The models are trained by minimizing a loss function suitable to MLC, typically binary cross-entropy (BCE, \cite{goodfellow2016deep}), whereby the predictions of the output neurons are independent. At inference time, a model outputs confidence scores for each class to be active, and thresholds are used to derive final binary predictions. One of the performance metrics most used is F-score, also called F1, which summarizes precision and recall values in a single number. In previous work, we compared parametric and non-parametric algorithms to search for optimal threshold values according to F1~\cite{cances2019evaluation}. In~\cite{kong2019sound}, the authors proposed to directly optimize F1 using gradient descent by computing F1 gradients numerically. These methods do bring consistent gains in F1 compared to non-optimized thresholds, but they have a strong limitation in terms of scalability to a large number of target classes. 

In this work, we propose a gradient descent based method that is fast and scalable thanks to the use of automatic differentiation tools. F1 is not differentiable because the inherent operation of thresholding is not. To circumvent this issue, instead of computing the gradients numerically as in~\cite{kong2019sound}, we propose to use Surrogate Gradient Learning (SGL,~\cite{Neftci2019a}), that gives the name of our proposed approach \textit{SGL\-Thresh}. To our knowledge, this method has never been used to optimize thresholds in multi-label classification, and for AT in particular. We report F1 results on three datasets, including AudioSet to assess scalability since AudioSet involves 527 target classes and about 20k examples.

\section{Problem statement\label{sec:problem}}



In this work, we aim at optimizing F1, a common metric to characterize performance with a single number. F1 is defined as the harmonic mean between precision and recall, and the larger, the better.
More precisely, we use the micro-averaged label F1 score, estimated at label level. The ground-truth and predictions are flattened out before computing micro-F1, as if there were a single class in the task at hand:
\[
    \text{label micro-F1} = \displaystyle\frac{2\sum_{n,l} y_l^{(n)}\hat{y}_l^{(n)}}{\sum_{n,l}y_l^{(n)}+\sum_{n,l}\hat{y}_l^{(n)}}
\]
where $y_l^{(n)}$ and $\hat{y}_l^{(n)}$ are binary values corresponding to the ground-truth and predictions for instance $n$ and label $l$. 
We chose to optimize micro-F1 because it has been used in Task 4 (SED task) of DCASE 2017 and 2019, two datasets on which we report results in this paper. 

The best threshold values depend on the target class of interest. We will illustrate this in the result section with a few examples. The central question of this work is how we could optimize the threshold values in order to obtain optimal F1 score values.






\section{Baseline threshold estimation methods}

We used two baseline methods for comparison purposes: defThresh, standing for ``default threshold va\-lue'', and dichoThresh, a dichotomic search algorithm.

\subsection{defThresh}

This method is the very baseline one, using a single default threshold value for all the classes, without any optimization procedure involved. We simply choose a value that gives the best compromise for all the classes. 

\subsection{dichoThresh: dichotomic search}

We proposed this method in previous work~\cite{cances2019evaluation}. This method optimizes class-dependent threshold values through a dichotomic search algorithm. The algorithm searches the best threshold values first within a coarse interval and coarse precision. It picks the ones that yield the best F1 score. From this first guess, small threshold decay values are randomly sampled from a Gaussian distribution centered on the preceding best threshold values. The complete process is repeated with an increased precision and a reduced search space. It stops when a maximum number of steps is reached.

The dichotomic search algorithm,  when compared to an exhaustive search of all the possible combinations, considerably reduces the time needed to reach efficient thresholds. However, the execution time is still dependent on the number of thresholds to tune and the amount of iterations for every step. 

\section{Gradient descent based methods}

In this section, we describe the two gradient-descent based methods used in this work: numThresh~\cite{kong2019sound} and SGL\-Thresh (ours). NumThresh and SGL\-Thresh share the same threshold optimization algorithm. Similarly to dichoThresh, we optimize class-dependent threshold values on a validation set (valid) and use them on an evaluation subset (eval) afterwards.

The thresholds are first initialized to a default value. Then, for a given number of epochs, the following steps are repeated: the network predictions (scores) are thresholded with the current threshold values with the Heaviside step function. The $\text{F1}$ objective function  and its gradients with respect to the thresholds are calculated. Then, the threshold values are updated with some optimization rule. In our experiments we used Adam~\cite{kingma2014adam}. At each epoch, the whole validation subset is used once to obtain robust gradient estimations (\textit{batch training}). Using minibatches led to worse performance in our experiments. 


The Heaviside step function is not differentiable at the origin and has a zero-valued gradient elsewhere. NumThresh and SGL\-Thresh differ in the way the gradient is approximated, as we shall describe below.

\subsection{numThresh: estimating F1 gradients numerically}

To handle F1 optimization in audio tagging, Kong and colleagues~\cite{kong2019weakly} approximate the gradients numerically:

\begin{equation}
    \nabla_{t}\text{F1}(t) = \displaystyle \frac{\text{F1}(t+\Delta t)-\text{F1}(t)}{\Delta t}
\end{equation}
where $t$ is a threshold, $\Delta t$ a small number (0.01 in practice). This computation is done for all the thresholds, which are then updated with the Adam rule~\cite{kingma2014adam}. The implementation provided by~\cite{kong2019sound} is very slow for large numbers of classes and examples, since all the directions are tested to maximize $\text{F1}$: for each threshold, a change in global $\text{F1}$ is tested if we increment $t$ by $\Delta t$ up to $10\Delta t$. Complexity is $\mathcal{O}(nC^2)$, where $C$ is the number of classes and $n$ is the number of examples. 


\subsection{SGL\-Thresh: Surrogate Gradient Learning (ours)}

SGL is a method to reconcile step functions, \textit{e.g.} the sign function and the Heaviside step function (HSF), with backpropagation.
It has been proposed to train ``binarized neural networks'', which are deep neural networks with weights and activations constrained to +1 or -1~\cite{Courbariaux2016}, and spiking neural networks, whose firing threshold is a HSF~\cite{Neftci2019a,zimmer2019technical}. 
In SGL\-Thresh, we use SGL to estimate the gradient $\nabla_{t}\text{F1}$ with the help of backpropagation and automatic differentiation.  

The idea behind of SGL is very simple: while the forward pass uses the HSF, the backward pass uses a ``surrogate gradient'', \textit{i.e.} the derivative of a differentiable function, a sigmoid function in our work: 

\begin{equation}
\Theta'(x) \approx \text{sig}_a'(x)= a \; \text{sig}_a(x) \; \text{sig}_a(-x)
\label{eq:surrogate}
\end{equation}
where $x=p-t$, $p$ being the network score predictions, and $t$ the class-dependent thresholds to be optimized. This approximation is used when computing the $\text{F1}$ gradients with respect to the thresholds:

\begin{equation}
\displaystyle \frac{\partial \text{F1}(t)}{\partial t} =  \frac{\partial \text{F1}}{\partial \hat{y}}\frac{\partial \hat{y}}{\partial t} \approx \frac{\partial \text{F1}}{\partial \hat{y}} \text{sig}_a'(p-t)
\label{eq:f1rule}
\end{equation}

In practice, SGL\-Thresh can be seen and is implemented as an activation function. In its forward mode, SGL\-Thresh is the HSF, and in its backward mode the partial derivatives of a sigmoid function. Our implementation in PyTorch is very fast since it benefits from PyTorch's vectorization and automatic differentiation capabilities to compute $\frac{\partial \text{F1}}{\partial \hat{y}}$.

\section{Audio tagging experiments \label{sec:applications}}

In this section, we report audio tagging experiments on three datasets: DCASE 2017~\cite{DCASE2017challenge} and DCASE 2019~\cite{serizel2018} task 4 and AudioSet eval~\cite{audioset}. DCASE refers to the Detection and Classification of Acoustic Scenes and Events yearly challenges, and its task 4 aims at detecting sound events in domestic environments. Regarding AudioSet, we used the official eval balanced subset.



\subsection{Datasets}

Table \ref{tab:datasets} shows the dataset characteristics: the number of target tags (classes), the number of samples used to optimize the thresholds (the \textit{val} subset), and the number of samples of the \textit{eval} subset, on which we apply the optimized thresholds for the final evaluation of the methods. We did not use the subsets on which the neural networks were trained. We hypothesize that optimizing thresholds on samples held out from the model training subset leads to more robust thresholds.

DCASE2017 and DCASE2019 are similar in terms of number of target classes: 17 and 10, respectively. AudioSet has many more classes (527 different tags), which is particularly interesting to test the scalability of the threshold optimizing methods regarding this aspect. For DCASE2017 and DCASE2019, we used the official val/eval splits, whereas for AudioSet, there is only an eval subset available. For that reason, in the case of AudioSet only, we ran a 3-fold cross-validation.



\begin{table}[htbp]
\caption{Dataset characteristics: number of target tags (classes), number of samples in val and eval subsets.}
\begin{center}
\begin{tabular}{lccc}
\toprule
name & classes & val & eval \\
\midrule
DCASE2017 & \hphantom{0}17 & \hphantom{00}488 & 1103 \\
DCASE2019 & \hphantom{0}10 & \hphantom{0}1122 & \hphantom{0}692 \\
AudioSet eval & 527 & 15278 & 5093 \\
\bottomrule
\end{tabular}
\label{tab:datasets}
\end{center}
\end{table}


\begin{table*}[htbp]
\caption{Micro-averaged F1 (\%) results in terms of on the evaluation subsets of DCASE\-2017, DCASE\-2019, AudioSet, between parentheses: on the valid subsets on which F1 was optimized. Optimization time is given for AudioSet.}
\begin{center}
\begin{tabular}{lcccc}
\toprule
 & DCASE\-2017 & DCASE\-2019 & AudioSet & Optimization time \\
 &           &           &      (3-fold CV)    & on AudioSet   \\
\midrule

defThresh & 59.0 (53.8) & 71.3 (70.1) & 50.7 (50.7) & \_ \\
dichoThresh & 61.4 (58.1) & 71.4 (71.7) & 53.5 (53.6) & 9 hours (CPU) \\ 
numThresh &  60.3 (58.7) & 72.2 (71.4) &  52.1 (53.6) & 5 days (CPU)\\ 
SGL\-Thresh (ours) & \textbf{62.0} (61.8) & \textbf{73.0} (72.9) & \textbf{54.9} (56.7) & 1.5 s (GPU), 3 s (CPU)\\

\bottomrule
\end{tabular}
\label{tab:resultsF1}
\end{center}
\vspace{-1.5mm}
\end{table*}




\subsection{Pre-trained neural networks}

Different state-of-the-art pretrained networks have been used for each of the three datasets. The acoustic features in all cases are log-Mel spectrograms, with slight differences in their extraction process though. 

For DCASE2017, we used the best performing model open-sourced from~\cite{kong2019sound}, called ``CNN-transformer-avg'' by the authors. It is comprised of four convolution blocks, each with two convolutional layers with $3\times 3$ kernel size, each layer followed by batch-normalization, Rectifier Linear Unit (ReLU) and  $2 \times 2 $ average pooling. The number of filters of the four blocks increases by a factor two: 64 filters for the first block, 128, 256 and 512 filters for the following blocks. A self-supervision module is placed after the last convolution block, hence the name invoking transformers~\cite{vaswani2017attention}. The total number of learnable parameters is 5.75 million parameters. 

For DCASE2019, we used the convolutional recurrent neural network of our own best submission to the challenge~\cite{Cances2019}, called ``model 1'' in~\cite{Cances2019}, with which we ranked fourth out of nineteen teams\footnote{http://dcase.community/challenge2019/task-sound-event-detection-in-domestic-environments-results}.  It is a small network of 165 thousand parameters, comprised of three convolution blocks followed by a bi-directional Gated Recurrent Unit layer (64 cells), a fully-connected layer of 10 units for localization and an audio tagging head with a 1024-unit fully-connected layer with ReLU and a 10-unit output layer with sigmoid activation functions. 

Finally, for the AudioSet experiments, we used the predictions made by the open-sourced model, with its available pretrained weights on AudioSet, called ``Cnn14\_mAP=0.431''~\cite{kong2019panns}. This model is comprised of six convolution blocks, each followed by a 20\% dropout layer, followed by the sum of a max- and a mean-pooling, a 2048-unit fully-connected layer with ReLU, and a 527-unit output layer with sigmoid activations. It totals 80.8 Million parameters.

\subsection{Results}



NumThresh thresholds were optimized with a learning rate of $1e$-2, with the Adam update rule, for 100 epochs, following~\cite{kong2019sound}. We used the same setting for SGL\-Thresh, but with a $1e$-3 learning rate. The initial thresholds were set to 0.3 for DCASE\-2017 and AudioSet, and 0.5 for DCASE\-2019. In SGL\-Thresh, we set the sigmoid slope $a$ to 50 in all the experiments. A large range of values for $a$, between 20 and 100, led to the same results. We ran the experiments on a machine with an Intel Xeon CPU E5-2623 and a GeForce GTX 1080 GPU.

In principle, the three methods should converge to the same threshold values, the ones that give the best F1 value on the valid subsets. Nevertheless, in practice, it is not the case, probably due to convergence issues. Hyperparameters, such as the threshold decay for dichoThresh and the learning rate for numThresh and SGLThresh, may not be optimal for each target class individually. Table~\ref{tab:resultsF1} compiles all the results in terms of micro-F1 values on the evaluation subset and within parentheses on the valid subset, on which the thresholds were optimized. The time needed to converge on AudioSet is given in the last column. As a first comment, we observe that our method, SGL\-Thresh, is by far the fastest method: only a few seconds to reach convergence either on GPU or CPU. DichoThresh needed about 9 hours and numThresh several days to converge, showing that the available implementation of numThresh is very inefficient when dealing with a large number of target classes and training examples. Globally speaking, the three optimization methods brought significant gains compared to using a default threshold (defThresh), and our proposed method, SGL\-Thresh, led to the best results on the eval subsets. On AudioSet in particular, the SGL\-Thresh gains are the largest ones, with 56.7\% and 54.9\% F1 values on the valid and eval subsets respectively. We ran 3-fold cross-validation on AudioSet, the F1 values are averaged on the three folds. Standard deviation was about 0.2\% in all cases.

DCASE\-2017 is the only corpus for which micro-F1 measures are available for reference in the literature. In 2017, the system best ranked in the challenge reached 55.6\%. Our model, which architecture is taken from~\cite{kong2019sound} and trained by ourselves, reached 59.0\% with defThresh and 60.3\% with numThresh, to be compared with 62.9\% and 64.6\% respectively, as reported by the authors. Although we did not reach these performance when training their model, we observe a similar relative gain in F1 brought by numThresh: 2.2\% relative. 

We reported results on micro-F1, but we also ran experiments optimizing macro-F1 scores, which we do not report for the sake of space. The results are similar to those of micro-F1: SGL\-Thresh outperformed the other methods in macro-F1 scores: 62.5\% compared to 53.4\% with defThresh on DCASE2017, for instance. 




\vspace{-1mm}
\section{Conclusions}

We proposed a new method to optimize F1 score, by searching for threshold values using surrogate gradient learning. We ran polyphonic audio tagging experiments on three benchmark datasets, and our method, called SGL\-Thresh, outperformed the use of a default threshold value by a large margin, but also two other optimization algorithms. Besides being accurate, SGL\-Thresh is also very fast thanks to automatic differentiation and vectorization. It scaled well on AudioSet, unlike the two other tested methods, and reached 54.9\% micro-F1 value on the eval subset. Future work will consist of end-to-end training models to optimize F1 instead of cross-entropy.

\section{Acknowledgements}

This work was partially supported by the French ANR agency within the LUDAU project (ANR-18-CE23-0005-01) and the French "Investing for the Future --- PIA3" AI Interdisciplinary Institute ANITI (Grant agreement ANR-19-PI3A-0004). We used HPC resources from CALMIP (Grant 2020-p20022).





\vfill\pagebreak


\bibliographystyle{IEEEbib}
\bibliography{strings,refs}

\begin{thebibliography}{10}

\bibitem{audioset}
Jort~F Gemmeke, Daniel~PW Ellis, Dylan Freedman, Aren Jansen, Wade Lawrence,
  R~Channing Moore, Manoj Plakal, and Marvin Ritter,
\newblock ``Audio set: An ontology and human-labeled dataset for audio
  events,''
\newblock in {\em Proc. ICASSP}, New Orleans, 2017, IEEE, pp. 776--780.

\bibitem{virtanen2018computational}
Tuomas Virtanen, Mark~D Plumbley, and Dan Ellis,
\newblock {\em Computational analysis of sound scenes and events},
\newblock Springer, 2018.

\bibitem{kong2019panns}
Qiuqiang Kong, Yin Cao, Turab Iqbal, Yuxuan Wang, Wenwu Wang, and Mark~D
  Plumbley,
\newblock ``{PANNS: Large-scale pretrained audio neural networks for audio
  pattern recognition},''
\newblock {\em arXiv preprint arXiv:1912.10211}, 2019.

\bibitem{goodfellow2016deep}
Ian Goodfellow, Yoshua Bengio, and Aaron Courville,
\newblock {\em Deep learning},
\newblock MIT press, 2016.

\bibitem{cances2019evaluation}
L{\'e}o Cances, Patrice Guyot, and Thomas Pellegrini,
\newblock ``Evaluation of post-processing algorithms for polyphonic sound event
  detection,''
\newblock in {\em 2019 IEEE Workshop on Applications of Signal Processing to
  Audio and Acoustics (WASPAA)}. IEEE, 2019, pp. 318--322.

\bibitem{kong2019sound}
Qiuqiang Kong, Yong Xu, Wenwu Wang, and Mark~D Plumbley,
\newblock ``Sound event detection of weakly labelled data with cnn-transformer
  and automatic threshold optimization,''
\newblock {\em arXiv preprint arXiv:1912.04761}, 2019.

\bibitem{Neftci2019a}
Emre~O. Neftci, Hesham Mostafa, and Friedemann Zenke,
\newblock ``{Surrogate Gradient Learning in Spiking Neural Networks},''
\newblock {\em IEEE Signal Processing Magazine}, vol. 36, no. October, pp.
  51--63, 2019.

\bibitem{kingma2014adam}
Diederik~P Kingma and Jimmy Ba,
\newblock ``Adam: A method for stochastic optimization,''
\newblock {\em arXiv preprint arXiv:1412.6980}, 2014.

\bibitem{kong2019weakly}
Qiuqiang Kong, Changsong Yu, Yong Xu, Turab Iqbal, Wenwu Wang, and Mark~D
  Plumbley,
\newblock ``Weakly labelled audioset tagging with attention neural networks,''
\newblock {\em IEEE/ACM Transactions on Audio, Speech, and Language
  Processing}, vol. 27, no. 11, pp. 1791--1802, 2019.

\bibitem{Courbariaux2016}
Matthieu Courbariaux, Itay Hubara, Daniel Soudry, Ran El-Yaniv, and Yoshua
  Bengio,
\newblock ``{Binarized Neural Networks: Training Deep Neural Networks with
  Weights and Activations Constrained to +1 or -1},''
\newblock {\em arXiv preprint arXiv:1602.02830}, 2016.

\bibitem{zimmer2019technical}
Romain Zimmer, Thomas Pellegrini, Srisht~Fateh Singh, and Timoth{\'e}e
  Masquelier,
\newblock ``Technical report: supervised training of convolutional spiking
  neural networks with pytorch,''
\newblock {\em arXiv preprint arXiv:1911.10124}, 2019.

\bibitem{DCASE2017challenge}
A.~Mesaros, T.~Heittola, A.~Diment, B.~Elizalde, A.~Shah, E.~Vincent, B.~Raj,
  and T.~Virtanen,
\newblock ``{DCASE} 2017 challenge setup: Tasks, datasets and baseline
  system,''
\newblock in {\em Proc. of the Detection and Classification of Acoustic Scenes
  and Events 2017 Workshop (DCASE2017)}, November 2017, pp. 85--92.

\bibitem{serizel2018}
Romain Serizel, Nicolas Turpault, Hamid Eghbal-Zadeh, and Ankit~Parag Shah,
\newblock ``Large-scale weakly labeled semi-supervised sound event detection in
  domestic environments,''
\newblock in {\em Proc. of the Detection and Classification of Acoustic Scenes
  and Events 2018 Workshop (DCASE2018)}, November 2018, pp. 19--23.

\bibitem{vaswani2017attention}
Ashish Vaswani, Noam Shazeer, Niki Parmar, Jakob Uszkoreit, Llion Jones,
  Aidan~N Gomez, {\L}ukasz Kaiser, and Illia Polosukhin,
\newblock ``Attention is all you need,''
\newblock in {\em Advances in neural information processing systems}, 2017, pp.
  5998--6008.

\bibitem{Cances2019}
Léo Cances, Thomas Pellegrini, and Patrice Guyot,
\newblock ``Multi-task learning and post processing optimization for sound
  event detection,''
\newblock Tech. {R}ep., IRIT, Université de Toulouse, CNRS, Toulouse, France,
  June 2019.

\end{thebibliography}

\end{document}